\documentclass{article}

\usepackage[preprint]{corl_2026} 
\usepackage{tikz}

\usepackage{booktabs}
\usepackage{multirow}
\usepackage{xcolor}
\usepackage{pifont}
\usepackage{amsmath}
\usepackage{amssymb}
\usepackage{graphicx}
\usepackage{siunitx}
\usepackage{comment}
\usepackage{enumitem}

\newcommand{\todocite}[1]{\textcolor{red}{REFERENCE}}
\newcommand{\change}[1]{\textcolor{blue}{#1}}

\title{ ContactFlow: A video action conditioning that transfers across embodiments}

\author{%
  Sami Azirar, Enrico Pallotta, Jan Nogga, Jürgen Gall, Sven Behnke, Hermann Blum\\[0.5em]
  University of Bonn, Lamarr Institute
}

\begin{document}
\maketitle


\begin{abstract}
World models offer a promising route toward robot planning by enabling agents to imagine and verify the consequences of actions before execution. However, current video-based world models often struggle to capture the physical constraints that govern manipulation, particularly contact. Further, their action conditioning is often constrained to specific embodiments such as parallel grippers. We propose \emph{Contact Flow}, an embodiment-agnostic action representation that encodes manipulation through the trajectory of 3D contact points between an actor and a target object. By discarding actor-specific appearance and kinematics, Contact Flow provides a shared conditioning signal for both human demonstrations and robotic execution. Therefore, we can train a large-scale video generative model on both human and robotic object interaction videos conditioned on Contact Flow, yielding a world model that predicts physically plausible manipulation outcomes.
We integrate this model into a propose-imagine-verify-act pipeline, where generated rollouts are assessed by a vision-language model before execution. Experiments on the DROID dataset and real-world tabletop manipulation tasks demonstrate that Contact Flow enables transfer between human demonstrations and different robotic embodiments.
\end{abstract}



\section{Introduction}\label{sec:Intro}

Teaching robots to interact with the physical world remains one of the central challenges of embodied intelligence. A promising direction is to equip robots with a \emph{world model}, a learned simulator of environment dynamics, that can be used to plan and verify actions before committing to execution. Such learned simulators have been used across a range of robotic applications:
for planning and control, by rolling out candidate action sequences in imagination and selecting those that reach the goal~\cite{du2023learning, du2024video}, for policy learning, by pretraining or distilling manipulation policies from generated visual dynamics~\cite{wu2024unleashing, liang2025video}, for data augmentation, by synthesizing novel demonstrations and experience to improve generalization~\cite{zhou2024robodreamer, jang2025dreamgen, liang2025dreamitate}, and for evaluation and verification, by acting as an action-conditioned simulator that scores or verifies behaviors before real-world execution~\cite{zhu2025irasim, wang2026interactive}.
Recent advances in video generation have made this increasingly tractable:
large-scale generative models~\cite{agarwal2025cosmos, wang2025wan} can now synthesize temporally coherent, photorealistic future frames conditioned on a variety of signals~\cite{jiang2025vace}. 
Yet physical faithfulness remains elusive~\cite{bansal2024videophy, motamed2025physicsiq}. Visually plausible videos need not 
respect the mechanics of contact and manipulation, but a world model that 
hallucinates physically impossible outcomes is of limited use for robot control.

We argue that one of the missing ingredients is an appropriate \emph{interaction 
representation}, a conditioning signal that encodes not what the scene looks 
like, but \emph{how} the agent plans to engage with it physically. Existing approaches 
fall into two families, each with a fundamental limitation. Joint-based 
representations~\cite{agarwal2025cosmos, pallotta2025egocontrol} are expressive but inherently embodiment-specific: a 
signal defined over a particular kinematic chain cannot transfer to a robot with 
a different morphology. Actor mask or silhouette representations~\cite{li2026mask2iv, yan2025open} avoid this to some extent but introduce a different bias: because they encode the full silhouette of 
the actor, they inadvertently place emphasis on the \emph{shape} of the actor (hand or end-effector) rather than on its interaction with the target object. Neither representation isolates what physically matters: the contact between the agent and the world.

This observation motivates a simple but consequential insight: physical 
manipulation is governed by contact. Regardless of the actor's morphology, an 
object's movement is only determined by where and how force is applied to it.
This can be captured as a compact, local, geometric signal that is agnostic to the embodiment producing it. 
We formalize this intuition as \textbf{Contact Flow}: the trajectory of 3D 
contact points between a hand or robotic grasp and the target object over time, 
projected into image space. Contact Flow is minimal by design, it discards 
everything about the actor except the locus of interaction, and it is 
naturally transferable across embodiments because it is defined entirely in terms 
of the object-centric contact geometry.

We use Contact Flow to condition a large-scale video generative model 
(Wan~\cite{wang2025wan}), training it on a heterogeneous mixture of human hand-object interaction (HOI) videos~\cite{zhao2025taste} and teleoperated robot demonstrations. The resulting world model generates manipulation rollouts and serves as a trajectory verifier: candidate action sequences are rendered as synthetic videos and evaluated by a vision-language model (VLM), only trajectories judged successful in simulation are passed to the robot for execution. 
Crucially, because Contact Flow abstracts away the actor, the very same signal is produced at inference directly from the robot's planned grasp on the target object: we read off the 3D contact points between the gripper and the object and project them into the camera, exactly as we extract them from human hands during training.
This enables zero-shot deployment in environments and on objects unseen during training.

We validate this pipeline on a held-out split of DROID~\cite{khazatsky2024droid}, on a suite of manipulation benchmarks unseen during training, and in real-world experiments on a fixed-arm manipulator observed from an exocentric camera, performing tabletop manipulation tasks in environments unseen during training.
Our results show that a model that learns interaction from human and robot demonstrations through a single embodiment-agnostic signal, can manipulate objects it has never encountered in environments unseen during training, a capability that we attribute directly to the embodiment-agnostic nature of Contact Flow. Our contributions are:
\begin{itemize}[nosep]
    \item \textbf{Contact Flow}, a novel action encoding shared by humans and robots: a compact, embodiment-agnostic representation based on the trajectory of 3D contact points between an actor and the target object, projected into image space.
    \item An \textbf{approach for conditioning a video world model on Contact Flow}, which we instantiate with both the ControlNet~\cite{zhang2023adding} and VACE~\cite{jiang2025vace} control-injection mechanisms and across multiple backbone scales.
    \item A \textbf{scalable data-processing pipeline} that extracts high-quality Contact Flow from heterogeneous human-demonstration and robot manipulation datasets as well as for zero-shot planning at inference.
\end{itemize}

\section{Related Work}
\label{sec:related_work}

\textbf{Video World Models for Robotics}
Recent work has explored generative video models as world models for robot planning, policy learning, data augmentation, and evaluation. UniPi~\cite{du2023learning} uses predicted videos as visual plans from which actions can be inferred. GR-1~\cite{wu2024unleashing} shows that large-scale video generative pretraining can improve language-conditioned manipulation policies. 
RoboDreamer~\cite{zhou2024robodreamer} learns compositional video world models for robot imagination, Dreamitate~\cite{liang2025dreamitate} transfers generated human demonstration videos to real-world visuomotor policies, and DreamGen~\cite{jang2025dreamgen} generates neural trajectories to improve generalization in robot learning. Recent works~\cite{zhu2025irasim,liang2025video,wang2026interactive} further treat video models as action-conditioned simulators or even policy representations.
A line of systems makes trajectories or explicit search the control interface for manipulation rollouts, such as depth-encoded trajectory-to-video generation~\cite{bai2025draw2act} and search-guided generative world models\cite{lin2025storm}.
We propose an embodiment-independent action representation to simulate manipulation trajectories for different agents rather than tying the world model to a specific robot or hand.

\textbf{Controllable video generation} provides the mechanism through which video world models can be steered toward task-relevant futures. Existing approaches differ mainly in the form of the conditioning signal. \cite{du2024video,jang2025dreamgen,zhen2025tesseract} rely on high-level semantic control, using language to specify goals or intermediate plans. 
A second line of work conditions generation on action or motion signals: \cite{zhu2025irasim,agarwal2025cosmos} use robot end-effector states, \cite{pallotta2025egocontrol} uses full-body human pose sequences. Finally, spatial control methods guide generation with image-aligned cues, including hand masks~\cite{akkerman2025interdyn}, hand-object mask trajectories~\cite{li2026mask2iv}, hand keypoints~\cite{zhao2025taste}, depth maps~\cite{wang2025precise}, optical-flow~\cite{jin2025flovd, koroglu2025onlyflow}, and motion-trajectory prompting~\cite{geng2025motion}.
These works show that the control signal is central to video generation for interaction. However, most existing signals are either high-level, embodiment-specific, or tied to the actor's visible morphology. In contrast, our work conditions generation on the contact geometry between actor and object, providing a local and embodiment-agnostic representation of the interaction.

\textbf{Hand-object interaction}
A large body of work studies hand-object interaction and the contact that mediates it. Several datasets and methods capture \emph{where} grasps touch objects: thermal contact maps in ContactDB~\cite{brahmbhatt2019contactdb}, paired hand-pose and object-contact annotations in ContactPose~\cite{brahmbhatt2020contactpose}, contact-driven grasp reconstruction in ContactOpt~\cite{grady2021contactopt}, and two-hand manipulation understanding in H2O~\cite{kwon2021h2o}. In robotics, contact geometry underpins grasp synthesis~\cite{sundermeyer2021contactgraspnet} and contact-rich manipulation planning~\cite{pang2023global}. These works detect, predict, or plan over contact as an analysis or control target. We use and repurpose estimated contact geometry as a generative conditioning signal.

\textbf{Cross-embodiments adaptation} is a key challenge for video world models. Regarding robot-to-robot transfer, Kinema4D~\cite{xu2026kinema4d} represents robot motion as 4D pointmaps encoding spatio-temporal robot occupancy, while BridgeV2W~\cite{chen2026bridgev2w} uses rendered embodiment masks to guide video generation across viewpoints, scenes, and robot platforms. These approaches reduce the dependence on raw robot actions, but still focus on the geometry of the acting embodiment. Other works use object-centric or flow-based interfaces, including flow as a cross-domain manipulation interface~\cite{xu2024flow}, any-point trajectory modeling~\cite{wen2024anypoint}, 3D object-flow~\cite{zhi2025flowaction}, and object-centric 3D motion fields~\cite{yin2025objectcentric}. These share our object-centric motivation but predict whole-object or surface motion, Contact Flow instead isolates the active contact interface that produces that motion.

\section{Method}
\label{sec:Method}
In the following we describe contact flow and how we use it to condition a video world model. We then describe how to extract contact flow from existing robotic and human demonstration datasets in order to train models. Finally, we explain how to build contact flow inputs at inference time.

\subsection{Contact Flow Encoding}
\label{subsec:contact_formulation}

Contact flow describes the surface interface where an agent makes contact with an object in the scene, as well as how that contact interface moves around in the scene over time. It captures where contact happens and how that contact region moves, but it does not include the object motion that happens afterward. For example, when someone turns a door handle, contact flow captures the motion of the handle surface where the fingers press and rotate it. It does not include the door swinging open afterward once the latch has released. Thus, it perfectly captures the active dynamics while excluding the passive dynamics. This is desirable because it focuses the learning signal on the interaction itself. It avoids training on signals that already contain the final outcome in the pixels, and it is not tied to a specific robot control interface. The same contact motion should produce the same representation, independent of whether a human hand or a robot gripper created it.
Formally, at each time step $t$ we represent contact flow as a set of contact points
\begin{equation}
\mathbf{C}_t = \big\{\, \mathbf{c}_t^{(i)} \,\big\}_{i=1}^{N_t}, \qquad
\mathbf{c}_t^{(i)} = \big( x,\, y,\, z,\; \Delta x,\, \Delta y,\, \Delta z,\; w \big) \in \mathbb{R}^{7},
\end{equation}
where $(x,y,z)$ is the 3D position of a contact point on the object surface in the camera frame, $(\Delta x, \Delta y, \Delta z)$ is its displacement to the following frame (the local contact motion, or ``flow''), and $w \in [0,1]$ is a confidence weight. Each point is projected into the image plane through the camera intrinsics $\mathbf{K}$ as $\mathbf{u}_t^{(i)} = \pi\big(\mathbf{K},\,(x,y,z)\big)$, and its seven attributes are written to the pixel $\mathbf{u}_t^{(i)}$ to form a sparse $7$-channel control frame, stacking these frames over time yields the spatiotemporal contact-flow video $\mathbf{C}_{1:T}$ that conditions the generator, rendered as a control video for the ControlNet~\cite{zhang2023adding} branch or encoded as a video condition for VACE~\cite{jiang2025vace}.

The confidence $w$ combines an estimate of how certain we are that a point is genuinely in contact with two geometric consistency cues. The base term is embodiment-specific: for robot data it is derived from the spatial closeness between the gripper and the object surface, whereas for human data it is the combined HaMeR hand-pose and HACO contact confidence (Sec.~\ref{subsec:data_proc}). We then modulate $w$ using the local spatio-temporal neighborhood: temporal alignment up-weights points whose displacement agrees in direction with that of their neighbors, and neighborhood density down-weights isolated points. Low-confidence points contribute less to both the conditioning signal and the training loss, which makes the representation robust to the noisy per-frame contact estimates produced by the extraction pipelines of Sec.~\ref{subsec:data_proc}.

\subsection{CF World Model}
\label{subsec:model}
We instantiate our Contact Flow world model with a latent video diffusion transformer following recent DiT-based video generators~\cite{wang2025wan}.
Given an input video $\mathbf{x}_{0:T} = \{x_0, x_1, \ldots, x_T\}$, we encode frames into a latent sequence $\mathbf{z}_{0:T} = \mathcal{E}(\mathbf{x}_{0:T})$, where $\mathcal{E}$ denotes the frozen video VAE encoder. 
The model is conditioned on the first observed frame $x_0$ and our contact flow signal $\mathbf{C}_{1:T}$, which specifies the intended interaction over the prediction horizon, learning a conditional generative model  \begin{equation} p_\theta(\mathbf{z}_{1:T} \mid z_0, \mathbf{C}_{1:T}). \end{equation}Training is performed with a flow-matching objective in latent space. We write the clean target latent sequence as $\mathbf{z}_{1:T} \sim q(\mathbf{z}_{1:T} \mid x_{1:T})$ and sample noise $\boldsymbol{\epsilon} \sim \mathcal{N}(0, I)$ of the same shape, the first-frame latent $z_0$ is kept clean as conditioning and is not noised.
For a sampled time $\tau \sim \mathcal{U}(0,1)$, we form the interpolant $\mathbf{u}_\tau = (1-\tau)\,\boldsymbol{\epsilon} + \tau\,\mathbf{z}_{1:T}, $ and train the DiT denoising network $v_\theta$ to predict the target velocity \begin{equation} \mathcal{L}_{\mathrm{FM}} = \mathbb{E}_{\mathbf{z}_{1:T}, \boldsymbol{\epsilon}, \tau} \left[ \left\| v_\theta(\mathbf{u}_\tau, \tau, z_0, \mathbf{C}_{1:T}) - (\mathbf{z}_{1:T} - \boldsymbol{\epsilon}) \right\|_2^2 \right]. \end{equation}
At inference time, the model starts from noise in latent space and integrates the learned velocity field conditioned on $z_0$ and $\mathbf{C}_{1:T}$ to generate future latents $\hat{\mathbf{z}}_{1:T}$, which are decoded into predicted frames $\hat{\mathbf{x}}_{1:T} = \mathcal{D}(\hat{\mathbf{z}}_{1:T})$. 
The goal is to condition future prediction on where and how the actor interacts with the object, rather than on the actor's appearance, hand shape, robot morphology, or embodiment-specific action space.

Control injection in diffusion-based generative models was popularized by ControlNet~\cite{zhang2023adding}, which introduces a trainable copy of the encoder blocks of a frozen U-Net, connected via zero-initialized convolution layers. 
Conditioning signals such as depth maps, pose skeletons, or edge maps are fed through this parallel branch, whose outputs are added back to the main network, enabling fine-grained structural control without modifying the pretrained backbone. 
More recently, VACE~\cite{jiang2025vace}, introduces a Video Condition Unit (VCU) that encodes control signals, reference frames, and binary spatiotemporal masks into a unified token representation, which is injected into a frozen Diffusion Transformer via lightweight Context Adapter blocks distributed across the network layers, enabling simultaneous and compositional control over both spatial structure and temporal dynamics.

In this work, we test our Contact Flow representation with both conditioning mechanisms to verify that its effectiveness is not tied to a specific control architecture. In the ControlNet-based variant, Contact Flow is rendered as a spatiotemporal control video and processed by the trainable control branch, which injects contact-aware features into the frozen video generation backbone. In the VACE-based variant, the same Contact Flow signal is encoded as an additional video condition and injected through the VCU and Context Adapters. Both variants use the same underlying representation: a sparse, object-centric trace of contact points projected into the image plane over time.

\subsection{Data Processing}
\label{subsec:data_proc}

We source training data from both human manipulation videos and teleoperated robot episodes. 
Human data provides scale and diversity of contact-rich interactions, while robot data offer metrically-accurate, embodiment-specific signals. Both streams are processed to recover the same downstream targets: object masks, hand/gripper pose, dense pointmaps, and contact regions.
For a video sequence $\lbrace x_t\rbrace_{t=0}^T$ of RGB frames $x_t$, we first acquire corresponding pointmaps $P_t$. Depending on the available sensing modalities, we recover these pointmaps either from stereo pairs, estimating metric depth with FoundationStereo~\cite{wen2025foundationstereo} and unprojecting it with the calibrated intrinsics, or directly from the RGB frames, optionally guided by noisy depth measurements, with MapAnything~\cite{keetha2026mapanything}.

\subsubsection{Contact Flow from Human Demonstration}

\textbf{Hand Mesh.} If not available in the dataset already, we first estimate hand poses in every frame using HaMeR~\cite{pavlakos2024reconstructing}. We identify occluded hand keypoints by their assigned confidence. For visible keypoints, we infer their 3D coordinates from the 2D output of HaMeR and look up the corresponding 3D points in the pointmap, because we find HaMeR's direct 3D estimation not consistent enough with the pointmap estimation. We then place the MANO model~\cite{romero2017mano} into the scene based on the estimated hand pose and shape parameters and fit it with a least-squares regression to the 3D coordinates of the visible keypoints.\\
\textbf{Object Mask.} We segment the manipulated object in each frame by prompting SAM3~\cite{carion2025sam3} with the target object's name. How that name is obtained depends on the cues the dataset provides: when a language description of the task is available, we extract candidate object nouns with spaCy~\cite{honnibal2020spacy}, when it is not, we instead query a vision-language model (Gemini~\cite{team2023gemini}) on the first frame $I_0$ to name the object being manipulated. If the dataset already ships object masks or meshes, we use these directly and skip detection. The resulting per-frame masks index into the pointmaps $P_t$ to recover the object's point cloud and provide the object support used during contact estimation.\\
\textbf{Contact Estimation.} We use HACO~\cite{jung2025haco} to estimate which parts of the hand mesh make contact with the object. It is only trained on right hands, which is why we feed it a left-right flip of our observations for left hands. We find that it is reliable at predicting which fingers etc. make contact, but it is not reliable in predicting whether the hand makes contact at all, also predicting random contact if the hand is meters away from any object. Therefore, we filter its output with a binary estimation of whether the hand is in contact with the object or not. For this, we measure whether any hand point is closer than $\delta_\textrm{contact dist}$ from the object, as well as a non-empty 2D overlap between the (dilated) hand mask and the object mask in the image. Contact for a single frame is estimated if both criteria are met. We further apply a temporal smoothing filter over this per-frame estimation.\\
\textbf{Dataset Specifics.} With the above steps, we process TasteRob~\cite{zhao2025taste}, TACO~\cite{liu2024taco}, and OakInk~\cite{zhan2024oakink2}. For TasteRob, we interpolate their provided object masks and depth to full frame rate and inpaint the holes in the depth map through nearest-neighbors. For TACO and OakInk, we process every camera view as a separate sample and take the provided hand and object meshes. We still run an additional quality check by passing the task annotation to GroundingDINO~\cite{liu2024groundingdino}, verifying that the predicted bounding box covers the reprojected object mesh. We skip samples where this check fails.

\subsubsection{Contact Flow from Robotic Data}
For robotic data we in general assume a calibrated, static, exocentric (i.e. looking at the arm, but not moving with the arm) camera, a URDF, and corresponding state vector available at every timestamp. Therefore, we can take a more geometric approach at contact estimation.\\
\textbf{Segmentation.} Given a text description of the task in the video, we enumerate candidate object nouns through spaCy~\cite{honnibal2020spacy}, and estimate 3D bounding boxes for each using WildDet3D~\cite{huang2026wilddet3d}, with ground-truth depth when available. From these candidates, we identify the object that the robot actually interacts with by checking for which candidate the robot's tool center point enters the 3D bounding box. We then take the pointmap $P_0$ and crop it with the 3D bounding box to retrieve the object's pointcloud. In addition, we prompt SAM3~\cite{carion2025sam3} with the selected object noun and $I_0$ to get a 2D object mask in the exocentric camera image. While in theory an accurate mask of the robot in the camera view can always be derived from the robot-to-camera calibration, robot state, and URDF rendering, we find that in practice small errors along this chain can easily lead to a few pixels offset. Therefore, we fine-tune a SAM3 model on the robot using the RoboEngine dataset~\cite{yuan2025roboengine} as well as 1000 hand-annotated images. \\ 
\textbf{Contact Estimation.} For contact, we distinguish robotic fingers that are between the camera and the object, and those that are, seen from the camera, behind the object. From the URDF and object pointcloud we know this distinction for each finger. For the rest of this paragraph, we assume a parallel gripper with one front and one back finger, but the method applies to any number of fingers.\\
For the front finger, we derive the overlap between the object mask and the robot mask and filter $P_0$ to this overlap region. The resulting points are points on the object that are in contact with the robot finger.
Since the back finger is not directly visible and also the object geometry is often incomplete on the backside, we have to take a different approach: We render the URDF into the image, if necessary adjusting its reprojection to the detected robot mask. We then take the rendering of the back finger and again check the overlap region with the object mask. The contact points are then taken from the rendering depth of the back finger within this overlap. Finally, we concatenate both sets to get the contact points for the given frame. For frames after gripper closing, we assume a rigid transform between the robot wrist and the last contact points, until the gripper is opened again. Point correspondences (“flow”) across frames are smoothed with a Hough-based
filter over a 3-frame window, which we empirically found to be the best operating point.We discard frames where the rendered robot and gripper disagree with the detected robot mask, and additionally filter with RobotInter~\cite{li2026robointer}, an interaction-prediction model that is not reliable enough to serve as a contact signal alone but lets us drop samples whose predicted interaction grossly disagrees with our estimate.

\subsubsection{Contact Flow at Inference}
At inference we require only an \emph{end-effector trajectory} together with a recovered model of the real-world scene, from the gripper geometry alone we synthesise the anticipated contact flow, so the conditioning signal is agnostic to whatever policy produced the trajectory. 
To populate the scene we need a metrically-accurate 3D model and pose of the target object together with a camera-frame pointmap: we assume access to at least one external stereo camera, to which we apply FoundationStereo~\cite{wen2025foundationstereo}.
As in training, the task is passed to a VLM (Gemini~\cite{team2023gemini}) that reduces it to an atomic instruction and proposes a bounding box for the target object, iteratively refining the box and optionally falling back to GroundingDINO~\cite{liu2024groundingdino}, and the selected object noun then prompts SAM3~\cite{carion2025sam3} for the object mask. We recover the object's 3D geometry by running SAM\,3D-Objects~\cite{sam3d2025} on the initial frame conditioned on the pointmap, as this typically leaves a pose error of several centimetres, we apply a staged refinement consisting of a closed-form PCA alignment with sign disambiguation followed by trimmed ICP against the observed object point cloud (compensating for the half-shell bias between the fully generated splat and the single-view visible surface), a differentiable render-and-compare step that optimises the global rigid pose and uniform scale of the Gaussian splat through \texttt{gsplat}~\cite{ye2024gsplat} differentiable rasterisation by minimising a composite loss of soft mask IoU, masked depth MSE, and per-pixel RGB cosine similarity, and a final per-Gaussian opacity pruning and colour-refinement step that sharpens silhouette and appearance, where each stage is accepted only if it improves all three strict render-space gatesmask IoU, depth inlier fraction, and RGB cosine similarity ($\geq 0.95$)ensuring the final splat faithfully reproduces the observed geometry and appearance under the calibrated camera. Finally, given the recovered scene, we apply the end-effector trajectory to the camera-frame pointmap through the gripper URDF, yielding the anticipated contact flow exactly as in our robotic-data extraction, and feed this contact flow to the video world model, which predicts how the real scene evolves under the proposed trajectory.


\section{Experiments}
\label{sec:experiments}
We investigate two questions: (Q1) Is Contact Flow an effective conditioning
signal, producing accurate, coherent predictions of robot manipulation? (Q2) Does the
world model predict real-world outcomes well enough to act as a zero-shot verifier of
proposed trajectories in unseen scenes?



%
\subsection{Experimental Setup}
%

\paragraph{Training data.}
We train on DROID,~\cite{khazatsky2024droid} Taste-ROB~\cite{zhao2025taste}, TACO~\cite{liu2024taco}, OakInk~\cite{zhan2024oakink2}, and LIBERO~\cite{liu2023libero}. We process this mix of human and robotic datasets as described in Section~\ref{sec:Method}.

\paragraph{Robot platform.}
Real-robot experiments are conducted on a Franka Panda fixed-arm manipulator with a single exocentric RGBD camera.
At deployment, we recover the scene once and instantiate the symbolic twin
(Sec.~\ref{sec:experiments}) in a single offline pass.
From the exocentric stereo pair we estimate metric depth with
FoundationStereo~\cite{wen2025foundationstereo}, prompt SAM3~\cite{carion2025sam3}
for the object mask, and fit a metric object mesh posed directly in the base
frame, refining its rigid pose and uniform scale until it is metrically
consistent with the observation through the differentiable procedure of
Sec.~\ref{subsec:data_proc}. With the twin in place, a $\pi_{0.5}$~\cite{physicalintelligence2025pi05}
vision-language-action policy rolls out a candidate end-effector trajectory
\emph{inside the twin}. We convert it to the anticipated contact flow, render
the predicted rollout with our contact-flow video world model, and a VLM
(Gemini~\cite{team2023gemini}) judges whether the task is solved.
Only then is the trajectory executed open-loop on the real robot.
\paragraph{Baselines and Metrics}
TesserAct~\cite{zhen2025tesseract} is
language-conditioned, taking only the task description, with no geometric control toward a
specific trajectory. CTRL-World~\cite{guo2025ctrl} conditions on low-dimensional action
embeddings, which are embodiment-specific and tied to the training robot's action space,
hence \emph{not} zero-shot. Kinema4D~\cite{xu2026kinema4d} conditions on a 4D pointmap of
the full robot, encoding the whole actor geometry rather than the contact locus alone.
Like ours it is zero-shot, and is our primary competitor. To test robustness independently
of backbone and control mechanism, we evaluate three configurations, Wan2.1-14B + VACE,
Wan2.2-5B + ControlNet, and Wan2.2-14B + ControlNet. Since Contact Flow conditions only on the object-side interface and never on the actor, we
compute the alignment metrics (PSNR, SSIM, LPIPS, DreamSim~\cite{fu2023dreamsim}) on the
agent-masked region, removing the agent (robot arm or human hand) from both prediction and
ground truth, with the same mask applied to the training loss. Masks come from RoboSeg
(robot) and SAM2 (human). FID and FVD are full-frame distribution metrics. All results use
$25$ held-out clips per dataset, each a $49$-frame window at $8$~FPS and $832\times480$.
DreamSim is our primary metric
For Q2 our primary measure is \emph{prediction accuracy}, whether the VLM forecast on the
imagined rollout matches real-robot execution for each trajectory $\pi_{0.5}$ proposes. A
trajectory succeeds only if the robot completes the task \emph{without unwanted
consequences} (e.g.\ knocking over or dropping objects), i.e.\ the target is grasped and
placed on the designated region within the episode.
\begin{table*}[t]
\centering
\caption{\textbf{Evaluation on DROID.}
All metrics are computed over $25$ held-out DROID clips after masking out the visible robot (via RoboSeg) in both the prediction and the ground truth. Arrows indicate the preferred direction of each metric.}
\label{tab:droid-eval}
\tiny
\setlength{\tabcolsep}{2.8pt}
\renewcommand{\arraystretch}{0.82}
\resizebox{0.92\textwidth}{!}{%
\begin{tabular}{@{}lllcccccc@{}}
\toprule
\textbf{Approach} & \textbf{Action Encoding} & \textbf{Backbone} & \textbf{DreamSim}$\downarrow$ & \textbf{FID}$\downarrow$ & \textbf{FVD}$\downarrow$ & \textbf{PSNR}$\uparrow$ & \textbf{SSIM}$\uparrow$ & \textbf{LPIPS}$\downarrow$ \\
\midrule
CF + VACE & 7ch & Wan 2.1 14B & 0.039 & 75.17 & 0.12 & 23.37 & 0.884 & 0.160 \\
CF + CTRL-Net & 7ch & Wan 2.2 5B & 0.036 & \textbf{40.41} & \textbf{0.03} & 23.28 & \textbf{0.904} & 0.165 \\
CF + CTRL-Net & 7ch & Wan 2.2 14B & \textbf{0.035} & 74.05 & 0.13 & \textbf{24.20} & 0.896 & 0.161 \\
\midrule
Kinema4D & Robot Pointmap & Wan 2.1 14B 4DNeX & 0.043 & 67.06 & 0.08 & 20.17 & 0.830 & 0.220 \\
CTRL-World & Action Emb. & SVD 1.5B & 0.059 & 72.90 & 0.07 & 22.44 & 0.800 & \textbf{0.090} \\
TesserAct & Language Cond. & CogVideo-X 5B & 0.106 & 164.54 & 0.34 & 13.96 & 0.615 & 0.396 \\
\bottomrule
\end{tabular}%
}
\end{table*}

\begin{figure*}[t]
\centering
\edef\rolloutheight{\the\dimexpr0.072\textwidth\relax}
\edef\simulatorheight{\the\dimexpr0.144\textwidth+4pt\relax}
\newcommand{\rolloutframe}[3]{%
  \includegraphics[trim=#2bp 0 #3bp 0,clip,height=\rolloutheight]{#1}}
\newcommand{\rolloutrow}[1]{%
  \hbox to \linewidth{\hfil
    \rolloutframe{#1}{0}{3888}%
    \rolloutframe{#1}{648}{3240}%
    \rolloutframe{#1}{1296}{2592}%
    \rolloutframe{#1}{2592}{1296}%
    \rolloutframe{#1}{3240}{648}%
    \rolloutframe{#1}{3888}{0}%
  \hfil}}
\newcommand{\rolloutframebright}[3]{%
  \includegraphics[trim=#2bp 0 #3bp 0,clip,height=\rolloutheight,decodearray={0 2.00 0 2.00 0 2.00}]{#1}}
\newcommand{\rolloutrowbright}[1]{%
  \hbox to \linewidth{\hfil
    \rolloutframebright{#1}{0}{3888}%
    \rolloutframebright{#1}{648}{3240}%
    \rolloutframebright{#1}{1296}{2592}%
    \rolloutframebright{#1}{2592}{1296}%
    \rolloutframebright{#1}{3240}{648}%
    \rolloutframebright{#1}{3888}{0}%
  \hfil}}
\begin{minipage}[t]{0.20\textwidth}
  \vspace{0pt}
  \centering
  \includegraphics[height=\simulatorheight]{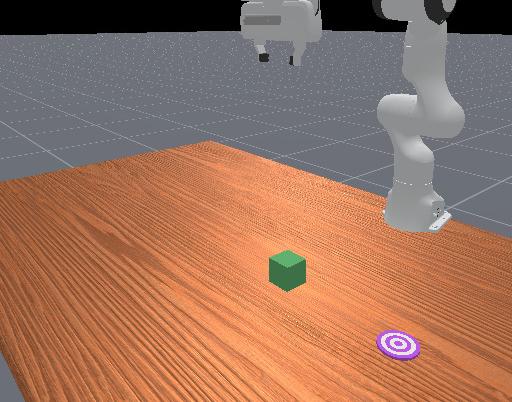}
\end{minipage}\hfill
\begin{minipage}[t]{0.76\textwidth}
  \vspace{0pt}
  \centering
  \vbox{%
    \rolloutrow{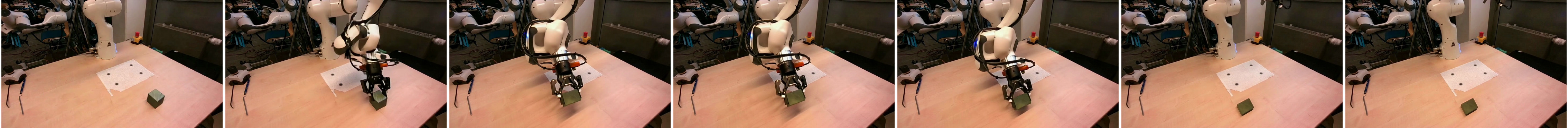}%
    \vskip4pt
    \rolloutrowbright{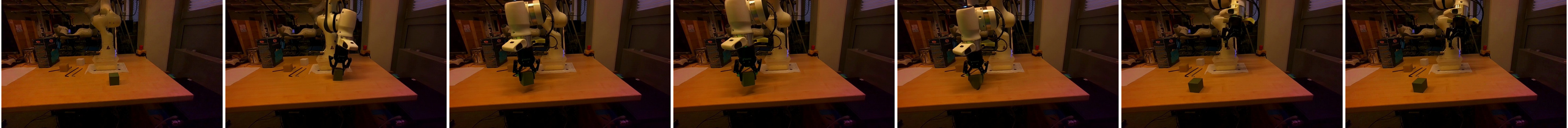}%
  }
\end{minipage}
\caption{\textbf{Closed-loop deployment with Contact Flow.}
On the left, a single RGBD view instantiates a symbolic twin simulator in which a
$\pi_{0.5}$ policy proposes an end-effector trajectory. On the right, our world model
imagines the rollout conditioned on the initial frame and Contact Flow from that trajectory,
showing the real rollout above and the simulated rollout below.}
\label{fig:deploy-rollout}
\end{figure*}

\paragraph{Results}
\textbf{(Q1)} We first evaluate Contact Flow as a conditioning signal for video prediction. On
held-out DROID scenes, each model receives the initial frame together with the
Contact Flow derived from the ground-truth end-effector trajectory and must
generate the full future video, which we score against the real recording
(Tab.~\ref{tab:droid-eval}). This measures prediction quality on held-out
scenes drawn from the training distribution, we probe generalisation to
entirely unseen embodiments, scenes, and rendering styles in the
cross-dataset evaluation below. We sample
these DROID frames at one third of the native rate, a stride of three, so a
single $49$-frame rollout forecasts roughly three times further into the future,
on the order of ten seconds of real interaction. We then extend
this evaluation beyond DROID to held-out splits of our human and robot training
data and to four datasets that are entirely unseen during training
(Tab.~\ref{tab:cross-dataset-transfer}), spanning new embodiments,
scenes, and rendering styles, which probes whether the same contact-flow interface
transfers across domains without any per-dataset adaptation. \textbf{(Q2)} We deploy the full pipeline on a Franka Panda across $10$ unseen tabletop pick-and-place scenarios. With the Wan~2.2 14B backbone, the world model forecasts the real-robot outcome correctly in $8/10$ cases. This verification step is what makes deployment work. Deploying the $\pi_{0.5}$ model directly into the real world dooes not enable any successful run despite it being a state-of-the-art VLA. The minimal twin lets the policy propose a motion, but it is too crude to certify that motion.

\begin{table*}[t]
\centering
\begin{minipage}[t]{0.49\textwidth}
\centering
\textbf{(a) Robot manipulation}\\[2pt]
\scriptsize
\setlength{\tabcolsep}{2pt}
\renewcommand{\arraystretch}{0.82}
\resizebox{\linewidth}{!}{%
\begin{tabular}{@{}llcccccc@{}}
\toprule
\textbf{Benchmark} & \textbf{Model} & \textbf{DreamSim}$\downarrow$ & \textbf{PSNR}$\uparrow$ & \textbf{SSIM}$\uparrow$ & \textbf{LPIPS}$\downarrow$ & \textbf{FID}$\downarrow$ & \textbf{FVD}$\downarrow$ \\
\midrule
\multirow{5}{*}{RLBench \textsc{ood}}
 & Kinema4D    & 0.051 & 26.11 & \textbf{0.872} & 0.176 & \textbf{81.79} & \textbf{0.12} \\
 & 5B          & 0.044 & 25.30 & 0.821 & 0.160 & 124.96 & 0.15 \\
 & 14B (mix)   & \textbf{0.040} & 25.44 & 0.826 & 0.164 & 106.66 & 0.14 \\
 & 14B (DROID) & 0.043 & \textbf{27.53} & 0.844 & \textbf{0.142} & 127.58 & 0.18 \\
 & 14B (human) & 0.061 & 25.23 & 0.817 & 0.175 & 126.16 & 0.18 \\
\midrule
\multirow{5}{*}{AgiBot \textsc{ood}}
 & Kinema4D    & 0.075 & 20.79 & \textbf{0.839} & \textbf{0.226} & 101.69 & 0.37 \\
 & 5B          & 0.066 & \textbf{21.10} & 0.770 & 0.261 & 107.51 & 0.44 \\
 & 14B (mix)   & \textbf{0.054} & 20.29 & 0.772 & 0.264 & \textbf{91.83} & \textbf{0.35} \\
 & 14B (DROID) & 0.082 & 19.97 & 0.801 & 0.233 & 107.92 & 0.44 \\
 & 14B (human) & 0.084 & 18.70 & 0.689 & 0.346 & 108.03 & 0.39 \\
\midrule
\multirow{5}{*}{GenieSim \textsc{ood}}
 & Kinema4D    & 0.085 & 21.21 & 0.858 & 0.253 & \textbf{92.99} & 0.49 \\
 & 5B          & \textbf{0.050} & \textbf{24.73} & \textbf{0.876} & \textbf{0.202} & 100.03 & 0.48 \\
 & 14B (mix)   & 0.063 & 22.45 & 0.856 & 0.222 & 101.10 & \textbf{0.44} \\
 & 14B (DROID) & 0.132 & 20.61 & 0.809 & 0.278 & 131.51 & 0.50 \\
 & 14B (human) & 0.075 & 22.31 & 0.823 & 0.271 & 108.15 & 0.45 \\
\bottomrule
\end{tabular}}
\end{minipage}\hfill
\begin{minipage}[t]{0.49\textwidth}
\centering
\textbf{(b) Human-hand manipulation}\\[2pt]
\scriptsize
\setlength{\tabcolsep}{2pt}
\renewcommand{\arraystretch}{0.82}
\resizebox{\linewidth}{!}{%
\begin{tabular}{@{}llcccccc@{}}
\toprule
\textbf{Benchmark} & \textbf{Model} & \textbf{DreamSim}$\downarrow$ & \textbf{PSNR}$\uparrow$ & \textbf{SSIM}$\uparrow$ & \textbf{LPIPS}$\downarrow$ & \textbf{FID}$\downarrow$ & \textbf{FVD}$\downarrow$ \\
\midrule
\multirow{5}{*}{TACO}
 & Kinema4D    & 0.070 & 23.18 & 0.803 & 0.243 & 88.29 & 0.29 \\
 & 5B          & 0.027 & 27.44 & 0.852 & 0.174 & 44.47 & 0.20 \\
 & 14B (mix)   & 0.024 & 27.22 & 0.857 & 0.168 & 39.50 & 0.20 \\
 & 14B (DROID) & 0.067 & 24.22 & 0.819 & 0.205 & 75.14 & 0.31 \\
 & 14B (human) & \textbf{0.021} & \textbf{28.28} & \textbf{0.869} & \textbf{0.158} & \textbf{36.46} & \textbf{0.18} \\
\midrule
\multirow{5}{*}{TASTE-Rob}
 & Kinema4D    & 0.107 & 20.68 & 0.783 & 0.221 & 175.95 & 0.52 \\
 & 5B          & \textbf{0.016} & 30.30 & 0.882 & 0.134 & 36.90 & \textbf{0.17} \\
 & 14B (mix)   & 0.019 & 29.11 & 0.888 & 0.134 & 36.34 & 0.21 \\
 & 14B (DROID) & 0.062 & 22.75 & 0.815 & 0.185 & 87.79 & 0.49 \\
 & 14B (human) & 0.016 & \textbf{30.66} & \textbf{0.899} & \textbf{0.124} & \textbf{32.62} & 0.20 \\
\midrule
\multirow{5}{*}{OakInk}
 & Kinema4D    & 0.074 & 22.08 & 0.797 & 0.266 & 105.66 & 0.35 \\
 & 5B          & 0.040 & 25.96 & 0.839 & 0.217 & 75.35 & 0.35 \\
 & 14B (mix)   & 0.032 & 26.09 & 0.852 & 0.207 & 58.10 & \textbf{0.33} \\
 & 14B (DROID) & 0.060 & 23.05 & 0.807 & 0.261 & 94.87 & 0.41 \\
 & 14B (human) & \textbf{0.029} & \textbf{26.77} & \textbf{0.854} & \textbf{0.199} & \textbf{57.10} & 0.33 \\
\midrule
\multirow{5}{*}{EgoDex$^{\dagger}$ \textsc{ood}}
 & Kinema4D    & 0.125 & \textbf{22.68} & \textbf{0.823} & \textbf{0.358} & 127.76 & \textbf{0.31} \\
 & 5B          & 0.127 & 22.11 & 0.818 & 0.372 & 128.33 & 0.36 \\
 & 14B (mix)   & 0.138 & 20.79 & 0.804 & 0.436 & 116.33 & 0.40 \\
 & 14B (DROID) & 0.218 & 19.79 & 0.770 & 0.410 & 167.14 & 0.43 \\
 & 14B (human) & \textbf{0.123} & 22.04 & 0.818 & 0.382 & \textbf{109.11} & 0.35 \\
\bottomrule
\end{tabular}}
\end{minipage}
\caption{\textbf{Cross-dataset evaluation.}
All metrics are computed over $25$ held-out clips per dataset after masking out the visible hand/robot in both the prediction and the ground truth.}
\label{tab:cross-dataset-transfer}
\end{table*}

We deploy the full pipeline on a Franka Panda across $10$ unseen tabletop pick-and-place scenarios. With the Wan~2.2 14B backbone, the world model forecasts the real-robot outcome correctly in $8/10$ cases. This verification step is what makes deployment work. Deploying the $\pi_{0.5}$ model directly into the real world dooes not enable any successful run despite it being a state-of-the-art VLA. The minimal twin lets the policy propose a motion, but it is too crude to certify that motion. 
\section{Limitations \& Conclusion}
We introduced \emph{Contact Flow}, an embodiment-agnostic action representation encoding manipulation as the trajectory of 3D contact points between actor and object. Conditioned on this signal, a single world model trained on mixed human and robot data predicts plausible outcomes and serves as a zero-shot verifier, transferring across embodiments, scenes, and objects unseen at training.\\
The world model captures interaction \emph{outcomes} more faithfully than the underlying \emph{physics}, which suffices for verification but could be made more physically grounded for contact-rich tasks.
The inference cost is still far away from real-time capabilities. 
In addition, contact quality depends on upstream calibration, metric depth, and the SAM3 masks, so the method assumes a calibrated exocentric camera with reliable depth.


\clearpage
\acknowledgments{If a paper is accepted, the final camera-ready version will (and probably should) include acknowledgments. All acknowledgments go at the end of the paper, including thanks to reviewers who gave useful comments, to colleagues who contributed to the ideas, and to funding agencies and corporate sponsors that provided financial support.}


\bibliography{references}  


\end{document}